\documentclass{article}

\PassOptionsToPackage{square,numbers}{natbib}
\usepackage[preprint]{neurips_2026}
\usepackage[utf8]{inputenc}
\usepackage[T1]{fontenc}
\usepackage{hyperref}
\usepackage{url}
\usepackage{microtype}
\AtBeginDocument{\definecolor{mydarkblue}{rgb}{0,0.08,0.45}%
\hypersetup{pdftitle={}, pdfauthor={Jaeyoon Jung, Suhwan Choi, Yunsung Lee, Youngjae Yu}, pdfborder=0 0 0, pdfpagemode=UseNone, colorlinks=true, linkcolor=mydarkblue, citecolor=mydarkblue, filecolor=mydarkblue, urlcolor=mydarkblue}}
\usepackage{amsmath,amssymb}
\usepackage{booktabs}
\usepackage{graphicx}
\usepackage{xcolor}
\usepackage{multirow}
\usepackage{makecell}
\usepackage{placeins}
\usepackage{tikz}
\usetikzlibrary{decorations.pathreplacing,positioning,arrows.meta}
\usepackage[font=small,labelfont=bf,skip=3pt]{caption}

\newcommand{\pp}{\textsc{PonderPounce}}
\newcommand{\ppwm}{\textsc{PonderPounce-WM}}
\newcommand{\ponder}{\textsc{Ponder}}
\newcommand{\pounce}{\textsc{Pounce}}
\newcommand{\pt}{\,pt}
\newcommand{\maintablestyle}{%
  \small
  \setlength{\tabcolsep}{4pt}%
  \renewcommand{\arraystretch}{1.08}%
}

\title{PonderPounce-WM: When Does Predicting the Future Help Context-Dependent Robot Control?}

\author{%
  Jaeyoon Jung\textsuperscript{1} \quad
  Suhwan Choi\textsuperscript{1,2} \quad
  Yunsung Lee\textsuperscript{1}\thanks{Co-corresponding authors.} \quad
  Youngjae Yu\textsuperscript{2}\footnotemark[2]\\[0.6em]
  {\normalfont\small
    \textsuperscript{1}MAUM.AI \qquad
    \textsuperscript{2}Seoul National University}
}

\begin{document}
\setcounter{footnote}{1}
\maketitle
\setcounter{footnote}{0}

\begin{abstract}
How does future prediction affect context-dependent robot control? We introduce \ppwm{}, extending \pp{} with three future tokens supervised to predict visual-feature changes. Its pretrained MLLM, \ponder{}, integrates history and demonstrations into continuous cognition for the action model, \pounce{}. We investigate token ordering and delivery to distinguish predictive supervision from direct future conditioning. On RoboMME, cognition-first ordering with cognition-only delivery improves success from 60.83\% to 67.42\%. With the same ordering, direct future-token delivery yields 64.00\% success. Probes show increased future information in cognition while the tested past-event attributes remain largely decodable. Across four real-robot tasks, \ppwm{} achieves 67.44\% mean success under asynchronous execution, compared with 60.98\% for \pp{}. These results suggest that future prediction can improve context-dependent control through auxiliary supervision without delivering future tokens to the action model.
\end{abstract}

\section{Introduction}
\label{sec:intro}

Context-dependent robot control requires retaining past events that are no longer observable and using demonstrations to interpret the current task. Predicting future observations may help learn such control representations, but the information needed to anticipate visual changes need not coincide with what an action model needs to act. Prior work combines observation history or explicit memory with future prediction for control~\citep{wu2024unleashing,shi2026memoryvla++,huang2026long}. Building on \pp{}'s use of a pretrained MLLM's native causal context as episode memory~\citep{choi2026ponderpounce}, we investigate how predictive supervision shapes this continuous context representation, called \emph{cognition}. We introduce \ppwm{}, which adds \emph{future tokens} supervised to predict visual-feature changes at multiple horizons, drawing on FLARE~\citep{zheng2025flare} and InternW0-$\Delta$~\citep{miao2026internw0}. This retains the MLLM context engine rather than introducing a video-generation backbone for pixel prediction.

\ppwm{} retains the dual-system architecture: \ponder{} integrates episode context, and \pounce{} generates actions conditioned on cognition. Our focus is how predictive supervision changes a representation already responsible for episode memory: whether it improves control, whether direct future conditioning adds a benefit, and whether past-event information remains decodable. We vary the causal ordering of cognition and future tokens and whether \pounce{} receives cognition alone or also the future tokens. On RoboMME~\citep{dai2026robomme}, future supervision with cognition-first ordering and cognition-only delivery improves success from 60.83\% to 67.42\%, while direct future-token delivery reduces the gain (Table~\ref{tab:main}). Future and past-event probes, together with inference-time token removal, examine how predictive learning changes cognition and its utility for control. Across four real-robot tasks under asynchronous execution, \ppwm{} achieves 67.44\% mean success versus 60.98\% for \pp{}, supporting real-world applicability (Table~\ref{tab:real}).

\section{\ppwm{}}
\label{sec:method}

We study world modeling as auxiliary prediction of future visual features from episode context, rather than action-conditioned rollout for planning.

\begin{figure}[t]
  \centering
  \includegraphics[width=0.96\textwidth]{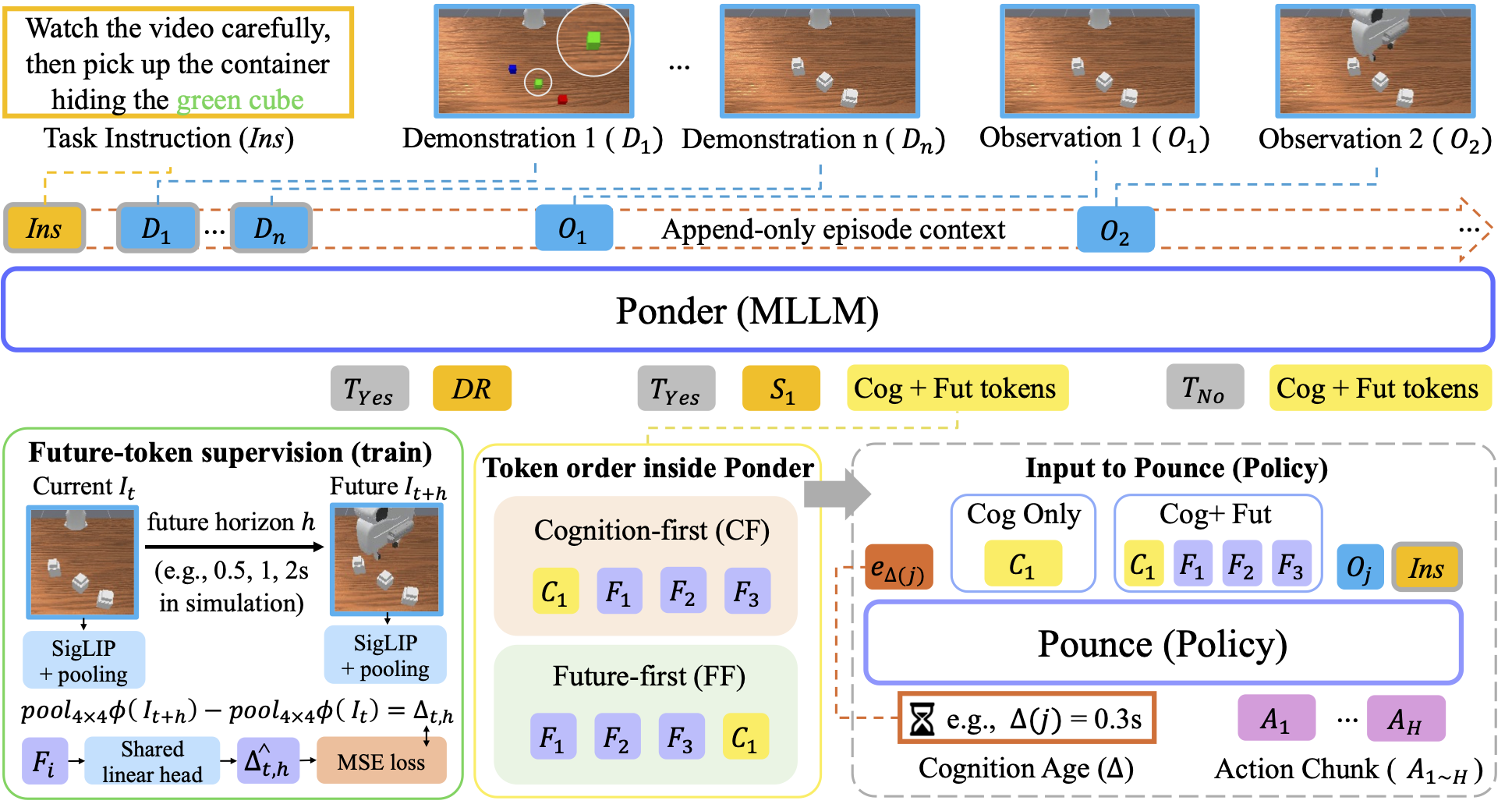}
  \caption{\textbf{\ppwm{} architecture.} \ponder{} appends one cognition token and three future tokens, with the latter trained to predict future visual-feature changes. The two axes are token order within \ponder{} (cog-first or fut-first) and delivery to \pounce{} (cognition alone or all four tokens).}
  \label{fig:arch}
\end{figure}

\paragraph{Base system.} \pp{}~\citep{choi2026ponderpounce} couples a pretrained MLLM, \ponder{}, which retains the task instruction, demonstrations, and execution history in its native causal context, with an action model, \pounce{}. At query $t$, it appends observation $\mathcal{O}_t$, optional text describing a new subgoal, and a learned cognition token. This token's final hidden state $\mathbf{c}_t\in\mathbb{R}^{H}$, with MLLM width $H$, is the continuous context representation passed to \pounce{}. At invocation $j$, \pounce{} combines its current observation, proprioception, and task with the prefix $[\mathbf{e}_{\Delta(j)},W_c\widetilde{\mathbf{c}}_j]$ to predict an action chunk. Here $W_c$ projects the newest ready cognition $\widetilde{\mathbf{c}}_j$ to the action model's width, and $\mathbf{e}_{\Delta(j)}$ encodes its age from the source observation. \pounce{} reuses this state until a newer \ponder{} output is ready. Both systems train end to end: action flow-matching loss propagates through the prefix, while grounding token cross-entropy enters through \ponder{}'s LM head. Action gradients into \ponder{} are scaled by 0.5, following the base recipe.

\paragraph{Future prediction.} We add three learned future tokens $F_i$, indexed by slot $i\in\{1,2,3\}$, giving $K{=}4$ tokens per query including cognition. Each slot predicts at horizon $h_i$, with $(h_1,h_2,h_3)=(0.5,1,2)$\,s in simulation and $(1,2,3)$\,s on the real robot (Appendix~\ref{app:real}). Let $\phi$ be \pounce{}'s frozen SigLIP encoder~\citep{zhai2023sigmoid} and $I_t$ the third-person image. The target is the feature change $\Delta_{t,h}=\mathrm{pool}_{4\times4}\phi(I_{t+h})-\mathrm{pool}_{4\times4}\phi(I_t)$, where pooling retains 16 spatial features. A shared linear head maps the hidden state $\mathbf{f}_{t,i}$ of slot $F_i$ to $\widehat{\Delta}_{t,h_i}\in\mathbb{R}^{16\times1152}$. The auxiliary loss $\mathcal{L}_{\mathrm{fut}}$ is the mean squared prediction error over valid query--horizon pairs, excluding targets beyond the episode end. We optimize $\mathcal{L}=\mathcal{L}_{\mathrm{fm}}+0.1\mathcal{L}_{\mathrm{ground}}+5\mathcal{L}_{\mathrm{fut}}$. Future-loss gradients update \ponder{}'s shared parameters and, when future tokens attend to cognition, flow through cognition as well. Thus predictive supervision can shape cognition without delivering future tokens to \pounce{}. The prediction head is not needed for action generation.

\paragraph{Token ordering and delivery.} The orders are $[\mathrm{cog},F_1,F_2,F_3]$ (cog-first) and $[F_1,F_2,F_3,\mathrm{cog}]$ (fut-first). Causal attention lets cognition read same-query future tokens only in fut-first. Both retain previous queries' tokens. Independently of order, \pounce{} receives cognition alone or all four hidden states through shared $W_c$, giving four configurations (Figure~\ref{fig:arch}).

\section{Experiments}
\label{sec:exp}

\paragraph{Setup.} RoboMME~\citep{dai2026robomme} has 16 tabletop tasks requiring counting, event memory, or following an in-context video demonstration, with 50 evaluation episodes each (800 per repeat). We use Qwen3.5-9B~\citep{qwen35blog} for \ponder{} and $\pi_{0.5}$~\citep{intelligence2025pi_} for \pounce{}. All checkpoints follow the same base \pp{} training recipe~\citep{choi2026ponderpounce}, apart from the slot and supervision settings under study (Appendix~\ref{app:training_setup}). RoboMME success rates (\%) average three closed-loop evaluations per checkpoint.

\paragraph{Main results.} Table~\ref{tab:main} shows that future supervision with cog-first ordering and cognition-only delivery reaches $67.42\%$, compared with $60.83\%$ for \pp{} ($K{=}1$) and $60.00\%$ for four slots without future supervision. The $K{=}4$ control disables future supervision and delivers all four slots to \pounce{}. All four serve as cognition tokens trained end to end through the action loss. Relative to $K{=}1$, gains concentrate in Counting and Imitation, with a smaller improvement in Permanence and little change in Reference. The largest gains occur on RouteStick and VideoUnmaskSwap, but MoveCube and PickHighlight decline by 10 and 9\pt{}, respectively (Appendix~\ref{app:scale}). Holding cog-first ordering fixed, direct delivery of future slots lowers success to $64.00\%$ ($-3.42$\pt{}). With fut-first ordering, the corresponding difference is $64.17\%$ to $63.17\%$ ($-1.00$\pt{}). Direct delivery thus yields a lower observed mean in both layouts, and the ordering difference is larger with cognition-only delivery. Paired episode-bootstrap 95\% intervals exclude zero for the cog-first delivery difference and the improvement over the $K{=}4$ no-future-loss control, but include zero for the fut-first delivery difference (Appendix~\ref{app:main_bootstrap}).

\begin{table}[!hbp]
  \vspace{-0.5em}
  \centering
  \caption{\textbf{RoboMME success rate.} Each family averages four tasks. FrameSamp+Modul and MemER results are from RoboMME~\citep{dai2026robomme}; the $K{=}1$ reference is from \citet{choi2026ponderpounce}. Best values per column are bold.}
  \label{tab:main}
  \maintablestyle
  \resizebox{\textwidth}{!}{%
  \begin{tabular}{lllrrrrr}
    \toprule
    Method & Slot order & \pounce{} reads & Counting & Permanence & Reference & Imitation & Average \\
    \midrule
    FrameSamp+Modul~\citep{dai2026robomme} & -- & -- & 65.22 & 25.11 & 36.33 & \textbf{51.39} & 44.51 \\
    MemER~\citep{sridhar2026scaling} & -- & -- & 48.83 & 53.16 & 38.00 & 29.50 & 42.38 \\
    \midrule
    \pp{} ($K{=}1$)~\citep{choi2026ponderpounce} & -- & 1 cog & 74.67 & 62.83 & 72.17 & 33.67 & 60.83 \\
    \pp{} ($K{=}4$) & -- & 4 cog & 74.33 & 61.33 & 66.50 & 37.83 & 60.00 \\
    \midrule
    \ppwm{} ($K{=}4$) & fut-first & cog + fut & 77.67 & 60.33 & 69.83 & 44.83 & 63.17 \\
    \ppwm{} ($K{=}4$) & fut-first & cog & 79.00 & 61.00 & \textbf{75.50} & 41.17 & 64.17 \\
    \ppwm{} ($K{=}4$) & cog-first & cog + fut & 77.50 & 59.33 & 73.00 & 46.17 & 64.00 \\
    \ppwm{} ($K{=}4$) & cog-first & cog & \textbf{84.83} & \textbf{68.83} & 72.00 & 44.00 & \textbf{67.42} \\
    \bottomrule
  \end{tabular}%
  }
  \vspace{-0.5em}
\end{table}
\FloatBarrier

\paragraph{Future information in cognition.} To test whether cognition carries more predictive information, we fit linear probes to future-feature changes on 192 held-out episodes. Table~\ref{tab:probe} shows that 1\,s explained variance rises from 0.25 to 0.31--0.34 and near-time discrimination from 0.63 to 0.67--0.69. Thus future information becomes more linearly decodable from cognition, even when future tokens are not delivered to \pounce{}. Removing execution history lowers explained variance at 1\,s by 0.10--0.14 across variants, indicating that context contributes to these predictions. Probe scores do not rank the variants in the same order as control success.

\begin{table}[!htbp]
  \vspace{-0.5em}
  \centering
  \maintablestyle
  \caption{\textbf{Future-feature read-out from cognition.} Scores are explained variance versus no change. ``Near'' is two-way accuracy. Definitions are in Appendix~\ref{app:compute}.}
  \label{tab:probe}
  \begin{tabular}{lllccccc}
    \toprule
    Method & Slot order & \pounce{} reads & 0.5\,s & 1\,s & 2\,s & 1\,s, no history & near \\
    \midrule
    \pp{} ($K{=}1$) & -- & 1 cog & 0.103 & 0.253 & 0.349 & 0.116 & 0.63 \\
    \pp{} ($K{=}4$) & -- & 4 cog & 0.098 & 0.239 & 0.329 & 0.115 & 0.62 \\
    \midrule
    \ppwm{} ($K{=}4$) & fut-first & cog + fut & 0.168 & 0.340 & 0.460 & 0.224 & 0.69 \\
    \ppwm{} ($K{=}4$) & fut-first & cog & 0.161 & 0.321 & 0.438 & 0.221 & 0.67 \\
    \ppwm{} ($K{=}4$) & cog-first & cog + fut & 0.158 & 0.313 & 0.409 & 0.207 & 0.67 \\
    \ppwm{} ($K{=}4$) & cog-first & cog & 0.172 & 0.336 & 0.436 & 0.209 & 0.69 \\
    \bottomrule
  \end{tabular}%
  \vspace{-0.5em}
\end{table}
\FloatBarrier

\paragraph{Past information in cognition.} We test whether past-event information remains decodable after future supervision. On 100 recorded held-out episodes per task, we probe hidden-container location, repetition ordinal, and per-colour counts of cubes already binned, using cognition at subgoal transitions (Appendix~\ref{app:memory}). Table~\ref{tab:memory} shows accuracies of 0.91--1.00 for container location, 0.99--1.00 for repetition ordinal, and 0.62--0.67 for BinFill colour counts, all above both current-frame features and current-observation cognition. Relative to $K{=}1$, future-supervised variants lose 1.5--4.5\pt{} in mean container accuracy, with bootstrap intervals excluding zero for two variants, while counting-group differences remain unresolved (Table~\ref{tab:memory_bootstrap}). The tested attributes remain largely decodable, but location read-outs show some degradation. Near-ceiling location and ordinal scores limit conclusions about broader memory preservation.

\begin{table}[!htbp]
  \vspace{-0.5em}
  \centering\footnotesize
  \setlength{\tabcolsep}{4pt}
  \caption{\textbf{Past-event read-out at subgoal transitions.} Held-out accuracy (0--1). The last row removes demonstrations and execution history from all six models and reports their accuracy ranges. Group means weight tasks equally. Probe details are in Appendix~\ref{app:memory}.}
  \label{tab:memory}
  \resizebox{\textwidth}{!}{%
  \begin{tabular}{lccccccc}
    \toprule
    & \multicolumn{2}{c}{hidden container (location)} & \multicolumn{3}{c}{counting (ordinal / colour counts)} & \multicolumn{2}{c}{group mean} \\
    \cmidrule(lr){2-3}\cmidrule(lr){4-6}\cmidrule(lr){7-8}
    Method & \makecell{Video\\Unmask} & \makecell{Button\\Unmask} & \makecell{Pick\\Xtimes} & \makecell{Swing\\Xtimes} & \makecell{Bin\\Fill} & container & counting \\
    \midrule
    chance & 0.27 & 0.27 & 0.33 & 0.33 & 0.32 & 0.27 & 0.33 \\
    mean position / majority class & 0.41 & 0.40 & 0.46 & 0.49 & 0.37 & 0.41 & 0.44 \\
    \midrule
    current frame (frozen SigLIP, no \ponder{}) & 0.41 & 0.41 & 0.80 & 0.66 & 0.55 & 0.41 & 0.67 \\
    \midrule
    \pp{} ($K{=}1$) & 1.00 & 0.99 & 1.00 & 1.00 & 0.63 & 0.99 & 0.88 \\
    \pp{} ($K{=}4$) & 0.99 & 0.98 & 1.00 & 1.00 & 0.62 & 0.98 & 0.87 \\
    \ppwm{} ($K{=}4$), fut-first / cog+fut & 0.98 & 0.97 & 1.00 & 1.00 & 0.67 & 0.97 & 0.89 \\
    \ppwm{} ($K{=}4$), fut-first / cog & 0.94 & 0.99 & 1.00 & 1.00 & 0.66 & 0.96 & 0.89 \\
    \ppwm{} ($K{=}4$), cog-first / cog+fut & 0.91 & 0.99 & 1.00 & 0.99 & 0.62 & 0.95 & 0.87 \\
    \ppwm{} ($K{=}4$), cog-first / cog & 1.00 & 0.96 & 1.00 & 1.00 & 0.65 & 0.98 & 0.88 \\
    \midrule
    \makecell[l]{All six models (range)\\Task + current observations only} & 0.42--0.42 & 0.30--0.38 & 0.75--0.80 & 0.60--0.65 & 0.49--0.55 & 0.36--0.40 & 0.62--0.66 \\
    \bottomrule
  \end{tabular}%
  }
  \vspace{-0.5em}
\end{table}
\FloatBarrier

\paragraph{Real-robot evaluation.} We evaluate \ppwm{} (cog-first / cog) on the same four RoboMME-inspired tasks and held-out scenes as the prior \pp{} evaluation~\citep{dai2026robomme,choi2026ponderpounce}, using wall-clock asynchronous execution (Appendix~\ref{app:real}). Table~\ref{tab:real} compares the new results with the reported baselines: mean success increases from 60.98\% to 67.44\%, with gains on Put Fruits, Track Cube, and Draw Pattern, but a decline on Repick Block (5/14 vs. 6/14). These results support real-world applicability, but separate evaluation sessions and differing cognition ages prevent isolating predictive supervision's effect. With 14--16 trials per task, task-level comparisons remain imprecise.

\begin{table}[!htbp]
  \vspace{-0.5em}
  \centering
  \maintablestyle
  \caption{\textbf{Real-robot success rates.} Entries show successes/trials (rounded \%). Average weights tasks equally. Prior results are from \citet{choi2026ponderpounce}.}
  \label{tab:real}
  \begin{tabular}{lccccc}
    \toprule
    Method & \makecell{Put Fruits\\(Counting)} & \makecell{Track Cube\\(Permanence)} & \makecell{Repick Block\\(Reference)} & \makecell{Draw Pattern\\(Imitation)} & Average \\
    \midrule
    $\pi_{0.5}$ & 9/14 (64\%) & 4/16 (25\%) & 0/14 (0\%) & 1/15 (7\%) & 23.99\% \\
    FrameSamp+Modul & 7/14 (50\%) & 5/16 (31\%) & 3/14 (21\%) & 9/15 (60\%) & 40.67\% \\
    \pp{} & 11/14 (79\%) & 10/16 (63\%) & \textbf{6/14 (43\%)} & 9/15 (60\%) & 60.98\% \\
    \midrule
    \makecell[l]{\ppwm{}\\(cog-first / cog)} & \textbf{12/14 (86\%)} & \textbf{12/16 (75\%)} & 5/14 (36\%) & \textbf{11/15 (73\%)} & \textbf{67.44\%} \\
    \bottomrule
  \end{tabular}%
  \vspace{-0.5em}
\end{table}
\FloatBarrier

\section{Conclusion}
\label{sec:conclusion}

\ppwm{} reaches 67.42\% success on RoboMME and 67.44\% across four real-robot tasks, versus 60.83\% and 60.98\% for \pp{}. Predictive supervision increases future-feature decodability while largely retaining tested past-event attributes, with some location degradation. Direct delivery yields lower observed success, while cog-first retains most gains without future tokens at inference (Appendix~\ref{app:inference_removal}). These findings connect predictive learning and control utility~\citep{yuan2026fast,gu2026right,yang2026mt} in an episode-memory setting. Comparisons cover one architecture and future targets from one encoder and camera view, without isolating future prediction from other auxiliary visual objectives or establishing the delivery-penalty mechanism and broader memory preservation. Within this scope, predictive supervision helps without delivering future tokens to the action model.

\clearpage
\begin{ack}
This work was supported by Institute of Information \& communications Technology Planning \& Evaluation (IITP) grant funded by the Korea government (MSIT) (No. RS-2026-25522885, Development of a World Foundation Model for Training and Deployment of Physical AI Systems).
\end{ack}

\bibliography{reference}

\clearpage
\appendix

\section{Per-task results}
\label{app:scale}

Table~\ref{tab:pertask} shows the task-level changes behind the family averages. Relative to the $K{=}1$ baseline, cog-first with cognition-only delivery gains on RouteStick ($+39$\pt{}), VideoUnmaskSwap ($+25$\pt{}), SwingXtimes ($+19$\pt{}), BinFill ($+11$\pt{}), VideoRepick ($+10$\pt{}), and StopCube ($+5$\pt{}), but loses on MoveCube ($-10$\pt{}) and PickHighlight ($-9$\pt{}), with smaller declines on ButtonUnmask ($-4$\pt{}), VideoUnmask ($-2$\pt{}), and VideoPlaceOrder ($-2$\pt{}) (differences rounded to whole percentage points). The largest gains are concentrated in RouteStick and VideoUnmaskSwap, so the overall improvement does not imply uniform gains across tasks. With cog-first ordering, delivering future slots to \pounce{} raises RouteStick further to 86\% but reduces performance on the swap and bin tasks. The fut-first/cognition-only variant retains 86\% on PickHighlight but also performs worse on both swap tasks than cog-first/cognition-only. Neither alternative preserves the full pattern of improvements.

\begin{table}[h]
  \centering\small
  \setlength{\tabcolsep}{2.5pt}
  \caption{\textbf{Per-task RoboMME success rates (\%).} Tasks are grouped by memory family, with abbreviations following RoboMME. Baseline results are from RoboMME~\citep{dai2026robomme}. Best values per column are bold.}
  \label{tab:pertask}
  \resizebox{\textwidth}{!}{%
  \begin{tabular}{lcccc|cccc|cccc|cccc|c}
    \toprule
    \multirow{2}{*}{Method} & \multicolumn{4}{c|}{Counting} & \multicolumn{4}{c|}{Permanence} & \multicolumn{4}{c|}{Reference} & \multicolumn{4}{c|}{Imitation} & \multirow{2}{*}{Average} \\
    \cmidrule(lr){2-5}\cmidrule(lr){6-9}\cmidrule(lr){10-13}\cmidrule(lr){14-17}
    & \makecell{Bin\\Fill} & \makecell{Pick\\Xtimes} & \makecell{Swing\\Xtimes} & \makecell{Stop\\Cube} & \makecell{Video\\Umsk} & \makecell{Button\\Umsk} & \makecell{Video\\UmskS} & \makecell{Button\\UmskS} & \makecell{Pick\\HighL} & \makecell{Video\\Repick} & \makecell{Video\\PlcBtn} & \makecell{Video\\PlcOrd} & \makecell{Move\\Cube} & \makecell{Insert\\Peg} & \makecell{Pattern\\Lock} & \makecell{Route\\Stick} & \\
    \midrule
    FrameSamp+Modul & 39.6 & 87.3 & \textbf{92.0} & 42.0 & 32.7 & 25.1 & 24.4 & 18.2 & 22.9 & 30.4 & 60.0 & 32.0 & 77.8 & 7.6 & \textbf{53.6} & 66.7 & 44.5 \\
    MemER & 56.7 & 79.3 & 59.3 & 0.0 & 81.3 & 72.0 & 38.0 & 21.3 & 70.7 & 25.3 & 30.0 & 26.0 & \textbf{82.7} & 6.7 & 16.7 & 12.0 & 42.4 \\
    \midrule
    \pp{} ($K{=}1$) & 50.7 & 93.3 & 72.0 & 82.7 & \textbf{94.7} & \textbf{97.3} & 25.3 & 34.0 & 85.3 & 40.7 & 83.3 & 79.3 & 81.3 & 6.0 & 12.7 & 34.7 & 60.8 \\
    \pp{} ($K{=}4$) & 56.7 & 97.3 & 76.0 & 67.3 & 92.0 & 90.7 & 27.3 & 35.3 & 68.0 & 32.7 & 81.3 & 84.0 & 78.7 & 8.7 & 18.0 & 46.0 & 60.0 \\
    \midrule
    \makecell[r]{\ppwm{}\\fut-first / cog+fut} & 54.7 & 97.3 & 84.7 & 74.0 & 92.0 & 93.3 & 27.3 & 28.7 & 78.7 & 42.0 & 78.0 & 80.7 & 70.7 & 12.0 & 18.7 & 78.0 & 63.2 \\
    \makecell[r]{\ppwm{}\\fut-first / cog} & 61.3 & 94.7 & 76.7 & 83.3 & 89.3 & 94.0 & 33.3 & 27.3 & \textbf{86.0} & 46.7 & 82.7 & \textbf{86.7} & 75.3 & 5.3 & 15.3 & 68.7 & 64.2 \\
    \makecell[r]{\ppwm{}\\cog-first / cog+fut} & 56.7 & 91.3 & 78.7 & 83.3 & 89.3 & \textbf{97.3} & 24.7 & 26.0 & 70.7 & \textbf{52.7} & 82.0 & \textbf{86.7} & 74.7 & 6.0 & 18.0 & \textbf{86.0} & 64.0 \\
    \makecell[r]{\ppwm{}\\cog-first / cog} & \textbf{62.0} & \textbf{98.7} & 90.7 & \textbf{88.0} & 92.7 & 93.3 & \textbf{50.7} & \textbf{38.7} & 76.0 & 50.7 & \textbf{84.0} & 77.3 & 71.3 & \textbf{13.3} & 17.3 & 74.0 & \textbf{67.4} \\
    \bottomrule
  \end{tabular}%
  }
\end{table}

\section{Paired bootstrap intervals for the main differences}
\label{app:main_bootstrap}

Every evaluation repeat of every checkpoint runs the same 800 evaluation slots (16 tasks $\times$ 50 episode indices). For each slot we take a model's success rate over its three repeats, form the per-slot difference between two models, and resample the 800 slots with replacement (2,000 draws, the same draws for every comparison) to obtain a percentile 95\% interval of the mean difference (Table~\ref{tab:main_bootstrap}). The intervals cover evaluation-slot sampling with repeat noise folded into the per-slot averages; they do not cover training seeds, and are not adjusted for multiple comparisons. The published $K{=}1$ reference is available only as per-task percentages, so it is not paired here; its three repeats span 59.62--62.25\% (Appendix~\ref{app:compute}). The improvement over the $K{=}4$ no-future-loss control, the cog-first delivery difference, and the ordering difference under cognition-only delivery exclude zero; the fut-first delivery difference and the ordering difference under all-slot delivery do not. An interval covering zero means the difference is not resolved by three repeats of 800 episodes, not that the variants perform the same.

\begin{table}[htbp]
  \centering\small
  \setlength{\tabcolsep}{5pt}
  \caption{\textbf{Paired episode-bootstrap differences in success rate.} Percentage points, A $-$ B, with percentile 95\% intervals over 2,000 resamples of the 800 evaluation slots; each slot's rate averages three repeats per model.}
  \label{tab:main_bootstrap}
  \resizebox{\textwidth}{!}{%
  \begin{tabular}{lllcc}
    \toprule
    comparison & A & B & mean & 95\% interval \\
    \midrule
    Best configuration vs. $K{=}4$ control & cog-first / cog & \pp{} ($K{=}4$) & $+7.42$ & $[+4.75,\,+10.21]$ \\
    Delivery (cog-first) & cog-first / cog & cog-first / cog + fut & $+3.42$ & $[+1.08,\,+5.92]$ \\
    Delivery (fut-first) & fut-first / cog & fut-first / cog + fut & $+1.00$ & $[-1.46,\,+3.50]$ \\
    Ordering (cog) & cog-first / cog & fut-first / cog & $+3.25$ & $[+0.71,\,+5.79]$ \\
    Ordering (cog + fut) & cog-first / cog + fut & fut-first / cog + fut & $+0.83$ & $[-1.54,\,+3.12]$ \\
    \bottomrule
  \end{tabular}}
\end{table}

\section{Past-event probe: protocol and uncertainty}
\label{app:memory}

\paragraph{Query selection.} We replay 100 held-out episodes per task through the serving path, retaining task instructions, demonstrations sampled at 1\,fps, and execution observations sampled at 1\,Hz. \ponder{} generates its own subgoal text, without ground-truth text in its context. We extract cognition at the first query at or after each annotated subgoal transition. Recorded observations are shared across checkpoints.

\paragraph{Targets and scoring.} We probe three past-event attributes:
\begin{itemize}
  \item \textbf{Hidden-container location.} VideoUnmask uses the first execution query, and ButtonUnmask the first query at the transition from button pressing to container picking. Ridge regression predicts the hidden container's image coordinate from cognition, using the recorded annotation as the target. A prediction is correct if its nearest container is the target. Candidate positions use a calibrated world-to-image affine map. Chance averages the inverse candidate count, and the constant reference predicts the mean target position.
  \item \textbf{Repetition ordinal.} PickXtimes and SwingXtimes queries are labelled by repetition index (first, second, or third). SwingXtimes includes both legs of each swing. A ridge read-out predicts one-hot labels and is scored by argmax.
  \item \textbf{BinFill colour counts.} At each pick-subgoal transition, the label is the tuple $(n_{\mathrm{red}},n_{\mathrm{green}},n_{\mathrm{blue}})$ of cubes already binned, identified from simulator actor identities, colours, and disappearance after earlier visibility, including cubes introduced during execution. Each tuple is a class, and accuracy requires an exact tuple match. There are 269 queries from 100 episodes. Chance is uniform over tuples allowed by the task's colour targets, excluding the completed state.
\end{itemize}
Labels are used only for probe fitting and scoring. Counting tasks additionally use a majority-class reference.

\paragraph{Read-out and references.} Ridge probes use three-fold cross-validation by episode, with penalties selected on an inner training split. Inputs are cognition hidden states, excluding future slots. The $K{=}4$ no-future-loss control averages its four cognition slots. The same procedure probes current-frame SigLIP features ($4{\times}4$ pooled) and each checkpoint's cognition given only task instructions and current observations, with no demonstrations or execution history. Group scores average tasks equally.

\paragraph{Uncertainty.} Table~\ref{tab:memory_bootstrap} reports paired percentile 95\% intervals from 2,000 episode-bootstrap resamples within each task. Resampling retains all queries of an episode and uses identical draws across checkpoints, with fixed out-of-fold predictions. Intervals cover episode sampling, not training seeds or probe fitting, and are not adjusted for multiple comparisons. Container accuracy declines are resolved for two future-supervised variants relative to $K{=}1$, while counting differences remain unresolved. Intervals containing zero do not establish equivalence.

\begin{table}[htbp]
  \centering\small
  \setlength{\tabcolsep}{5pt}
  \caption{\textbf{Paired episode-bootstrap differences in past-event read-out.} Percentage points with 95\% intervals. Positive favours the future-supervised variant.}
  \label{tab:memory_bootstrap}
  \resizebox{\textwidth}{!}{%
  \begin{tabular}{lcccc}
    \toprule
    & \multicolumn{2}{c}{vs.\ \pp{} ($K{=}1$)} & \multicolumn{2}{c}{vs.\ \pp{} ($K{=}4$)} \\
    \cmidrule(lr){2-3}\cmidrule(lr){4-5}
    future-supervised variant & hidden container & counting & hidden container & counting \\
    \midrule
    fut-first / cog+fut & $-2.0\;[-4.5,\,0.5]$ & $+1.2\;[-0.7,\,3.0]$ & $-1.0\;[-4.0,\,2.0]$ & $+1.5\;[-0.4,\,3.3]$ \\
    fut-first / cog & $-3.0\;[-6.0,\,-0.5]$ & $+0.8\;[-1.0,\,2.5]$ & $-2.0\;[-5.0,\,0.5]$ & $+1.1\;[-0.6,\,2.8]$ \\
    cog-first / cog+fut & $-4.5\;[-8.0,\,-1.5]$ & $-0.8\;[-2.2,\,0.6]$ & $-3.5\;[-6.5,\,-0.5]$ & $-0.5\;[-2.1,\,1.1]$ \\
    cog-first / cog & $-1.5\;[-4.0,\,0.5]$ & $+0.5\;[-1.1,\,2.2]$ & $-0.5\;[-3.0,\,2.0]$ & $+0.9\;[-0.9,\,2.5]$ \\
    \bottomrule
  \end{tabular}%
  }
\end{table}

\section{Future-token removal at inference}
\label{app:inference_removal}

Following work that separates predictive training from test-time future generation~\citep{yuan2026fast,yang2026mt}, we omit all three future tokens from \ponder{} without retraining. Both checkpoints were trained with cognition-only delivery, so \pounce{}'s input remains one cognition slot. Success changes from 67.42\% to 66.17\% for cog-first and from 64.17\% to 8.75\% for fut-first (Table~\ref{tab:inference}). Cog-first retains most of its gain, but removal changes the accumulated context and token positions in both layouts, and additionally removes same-query future-token access in fut-first. The intervention therefore does not isolate these sources of dependence.

\begin{table}[!htbp]
  \centering
  \maintablestyle
  \caption{\textbf{Removing future slots at inference.} RoboMME success (\%) without retraining. The inference column lists tokens computed by \ponder{}. \pounce{} always receives cognition only.}
  \label{tab:inference}
  \begin{tabular}{llrrrrr}
    \toprule
    Training order & \ponder{} inference & Counting & Permanence & Reference & Imitation & Average \\
    \midrule
    fut-first & Cognition + future & 79.00 & 61.00 & 75.50 & 41.17 & 64.17 \\
    fut-first & Cognition only & 6.83 & 7.67 & 7.83 & 12.67 & 8.75 \\
    \midrule
    cog-first & Cognition + future & 84.83 & 68.83 & 72.00 & 44.00 & 67.42 \\
    cog-first & Cognition only & 80.00 & 68.17 & 72.50 & 44.00 & 66.17 \\
    \bottomrule
  \end{tabular}%
\end{table}

\section{Horizon-matched future-slot probes}
\label{app:future_slot_probe}

For each frozen checkpoint, a separate ridge probe predicts $\Delta_{t,h_i}$ from only future slot $F_i$, with $(h_1,h_2,h_3)=(0.5,1,2)$\,s. Slots are not concatenated, and the probes are fitted independently of the trained prediction head. We use the cognition probes' 192 episodes, query selection, targets, and cross-validation protocol (Appendix~\ref{app:compute}).

Table~\ref{tab:future_slot_probe} gives explained variance of approximately 0.39, 0.52, and 0.59 across the three horizons, exceeding the corresponding cognition read-outs. Scores differ by at most 0.004 across configurations. Cross-horizon read-out degrades with distance from a slot's trained horizon, consistent with horizon-specific predictive information.

\begin{table}[htbp]
  \centering
  \maintablestyle
  \caption{\textbf{Horizon-matched future-slot read-out.} Held-out explained variance versus no change, using only the future slot for each target horizon.}
  \label{tab:future_slot_probe}
  \resizebox{\textwidth}{!}{%
  \begin{tabular}{lllccc}
    \toprule
    Method & Slot order & \pounce{} reads & $F_1\!\to\!0.5$\,s & $F_2\!\to\!1$\,s & $F_3\!\to\!2$\,s \\
    \midrule
    \ppwm{} ($K{=}4$) & fut-first & cog + fut & 0.390 & 0.521 & 0.592 \\
    \ppwm{} ($K{=}4$) & fut-first & cog & 0.390 & 0.521 & 0.592 \\
    \ppwm{} ($K{=}4$) & cog-first & cog + fut & 0.388 & 0.517 & 0.592 \\
    \ppwm{} ($K{=}4$) & cog-first & cog & 0.389 & 0.520 & 0.590 \\
    \bottomrule
  \end{tabular}%
  }
\end{table}

\section{Real-robot experiments}
\label{app:real}

\paragraph{Platform and tasks.} We follow the prior \pp{} setup~\citep{choi2026ponderpounce}, using the right arm of a Trossen ALOHA Stationary AI Kit with external and wrist RGB views and proprioception. The four RoboMME-inspired tasks~\citep{dai2026robomme} are PutFruits (transfer a specified number of fruits and press a stop button), TrackCube (select the cup hiding the specified-colour cube), RepickBlock (pick blocks in the demonstrated order), and DrawPattern (reproduce a demonstrated path). The latter three use pre-execution demonstration videos.

\paragraph{Data and models.} The 710 teleoperated episodes have human-annotated subgoals but no demonstration reasoning. Training-episode/held-out-scene counts are 104/14, 195/16, 206/14, and 205/15 for the four tasks above. Evaluation scenes and demonstration clips are held out from training. The first three rows of Table~\ref{tab:real} are prior results~\citep{choi2026ponderpounce}. We train \ppwm{} (cog-first / cog) with the simulation recipe in Appendix~\ref{app:training_setup}, using future horizons of 1, 2, and 3\,s, subgoal-only grounding, 30-step action chunks, and a 16K training context.

\paragraph{Evaluation and timing.} \ponder{} and \pounce{} target 1\,Hz on separate H100 GPUs, with 30\,Hz action playback and a 131K serving context cap. \pounce{} uses the latest ready cognition and its age. A human judges success or failure under a two-minute limit. Each held-out scene contributes one trial, and the reported average weights the four task rates equally. Table~\ref{tab:real_timing} pools server-side timing samples across 59 trials per model. \ponder{} compute time includes transition-triggered text decoding. Cognition age runs from its source observation to the \pounce{} invocation using it, and action observation age from the \pounce{} observation to action delivery. Clock phase can affect cognition age, so these deployment measurements do not isolate future-token overhead.

\begin{table}[htbp]
  \centering\small
  \setlength{\tabcolsep}{5pt}
  \caption{\textbf{Real-robot asynchronous deployment timing.} Milliseconds, with percentiles pooled over samples from 59 trials per model. \pp{} measurements use the prior deployment's recordings~\citep{choi2026ponderpounce} and the same aggregation procedure. Cognition age excludes post-freeze samples.}
  \label{tab:real_timing}
  \resizebox{\textwidth}{!}{%
  \begin{tabular}{llrrrr}
    \toprule
    & & \multicolumn{2}{c}{\pp{}} & \multicolumn{2}{c}{\ppwm{} (cog-first / cog)} \\
    \cmidrule(lr){3-4}\cmidrule(lr){5-6}
    System & Metric & Median & p95 & Median & p95 \\
    \midrule
    System~1 (\pounce{}) & Invocation interval & 1,000 & 1,001 & 1,000 & 1,001 \\
     & Compute time & 181 & 241 & 204 & 254 \\
     & Non-null cognition age & 1,797 & 2,003 & 1,022 & 1,039 \\
    \midrule
    System~2 (\ponder{}) & Invocation interval & 1,000 & 1,001 & 1,000 & 1,002 \\
     & Compute time & 187 & 537 & 223 & 584 \\
    \midrule
    Action delivery & Observation age & 776 & 1,223 & 777 & 1,224 \\
    \bottomrule
  \end{tabular}%
  }
\end{table}

\section{Training configuration}
\label{app:training_setup}

We follow the \pp{} recipe~\citep{choi2026ponderpounce} on RoboMME's 1,587-episode base split with grounded subgoals and demonstration reasoning. Training uses 4,320 steps, global batch 32 on 32 H100 GPUs, learning rate $2{\times}10^{-5}$ with 100 warm-up steps followed by a constant schedule, gradient clipping at 1.0, and ZeRO-2. Both backbones use public pretrained weights, with newly initialized carrier tokens and projections. Training samples \pounce{} invocations every 100\,ms and \ponder{} queries every 1\,s with a simulated 300\,ms delay. All four future-supervised configurations share the data, objectives, schedule, and training seed. The $K{=}4$ control sets the future-loss weight to zero and delivers all four slots to \pounce{}. Real-robot differences are listed in Appendix~\ref{app:real}.

\section{Compute and evaluation details}
\label{app:compute}

\paragraph{Simulation evaluation.} The $K{=}1$ reference uses the prior three-repeat result of 60.83\% (range 59.62--62.25\%) under the same serving protocol~\citep{choi2026ponderpounce}. Both systems run at 1\,Hz in simulated time with a 300\,ms cognition delay. \pounce{} uses 10 flow-matching steps and plays 20-step chunks at 20\,Hz, starting after the first cognition is ready. Each checkpoint is evaluated three times on the same 50 episodes per task, capped at 1,300 environment steps.

\paragraph{Future-probe protocol and metrics.} Probes use 192 held-out episodes and 2,752 queries, excluding targets beyond episode end. Three-fold cross-validation splits by episode, with ridge penalties selected on an inner training split. Explained variance is $1-\operatorname{MSE}(\widehat{\Delta},\Delta)/\operatorname{MSE}(0,\Delta)$, averaged over held-out folds, with zero matching no-change prediction. The \emph{1\,s, no history} condition removes earlier execution observations, subgoals, and carrier tokens but retains the task, demonstrations, current observations, and current-query grounding text. \emph{Near} is two-way accuracy comparing the predicted 1\,s delta with the true target and a same-episode distractor offset by one or two queries. The no-change predictor scores 0.51.

\end{document}